\documentclass[letterpaper, 10 pt, journal, twoside]{IEEEtran}  
\IEEEoverridecommandlockouts

\usepackage[utf8]{inputenc}
\usepackage{graphicx}
\usepackage{hyperref}
\usepackage{cleveref}
\Crefname{figure}{Fig.}{Figs.}
\Crefname{section}{Sec.}{Secs.}
\Crefname{table}{Tab.}{Tabs.}
\Crefname{equation}{Eqn.}{Eqns.}

\usepackage[per-mode=symbol]{siunitx}            
\usepackage[dvipsnames]{xcolor}
\definecolor{forestgreen}{RGB}{34 139 34}
\definecolor{darkorchid}{RGB}{153 50 204}
\definecolor{darkorange}{RGB}{255 140 0}

\title{Diaphragm Ankle Actuation for Efficient \\ Series Elastic Legged Robot Hopping}

\author{Marco Bolignari$^{1,2}$%
, An Mo$^{3}$%
, Marco Fontana$^{2}$ and Alexander Badri-Spr\"owitz$^{3}$
\thanks{This work was supported by Department of Excellence grant of Scuola Superiore Sant’Anna, the International Max Planck Research School for Intelligent Systems, the China Scholarship Council, and the Max Planck Society. Gefördert durch die Deutsche Forschungsgemeinschaft (DFG)-449912641.}
\thanks{$^{1}$ Department of Industrial Engineering, University of Trento, 38123 Trento, Italy. \tt \footnotesize {[marco.bolignari]}@unitn.it}
\thanks{$^{2}$ Institute of Mechanical Intelligence, Scuola Superiore Sant'Anna, 56127 Pisa, Italy. \tt \footnotesize {[marco.fontana]@santannapisa.it}}
\thanks{$^{3}$ Dynamic Locomotion Group, Max Planck Institute for Intelligent Systems, 70569 Stuttgart, Germany. \tt \footnotesize {[mo]}{[sprowitz]}@is.mpg.de}
}

\begin{document}
\maketitle

\begin{abstract}
Robots need lightweight legs for agile locomotion, and intrinsic series elastic compliance has proven to be a major ingredient for energy-efficient locomotion and robust locomotion control. Animals' anatomy and locomotion capabilities emphasize the importance of that lightweight legs and integrated, compact, series elastically actuated for distal leg joints. But unlike robots, animals achieve series elastic actuation by their muscle-tendon units. So far no designs are available that feature all characteristics of a perfect distal legged locomotion actuator; a low-weight and low-inertia design, with high mechanical efficiency, no stick and sliding friction, low mechanical complexity, high-power output while being easy to mount. Ideally, such an actuator can be controlled directly and without mechanical cross-coupling, for example remotely. With this goal in mind, we propose a low-friction, lightweight Series ELastic Diaphragm distal Actuator (SELDA) which meets many, although not all, of the above requirements. We develop, implement, and characterize a bioinspired robot leg that features a SELDA-actuated foot segment. We compare two leg configurations controlled by a central pattern generator that both feature agile forward hopping. By tuning SELDA's activation timing, we effectively adjust the robot's hopping height by 11\% and its forward velocity by 14\%, even with comparatively low power injection to the distal joint.
\end{abstract}

\begin{figure}[t]
    \centering
    \includegraphics[width=0.8\linewidth]{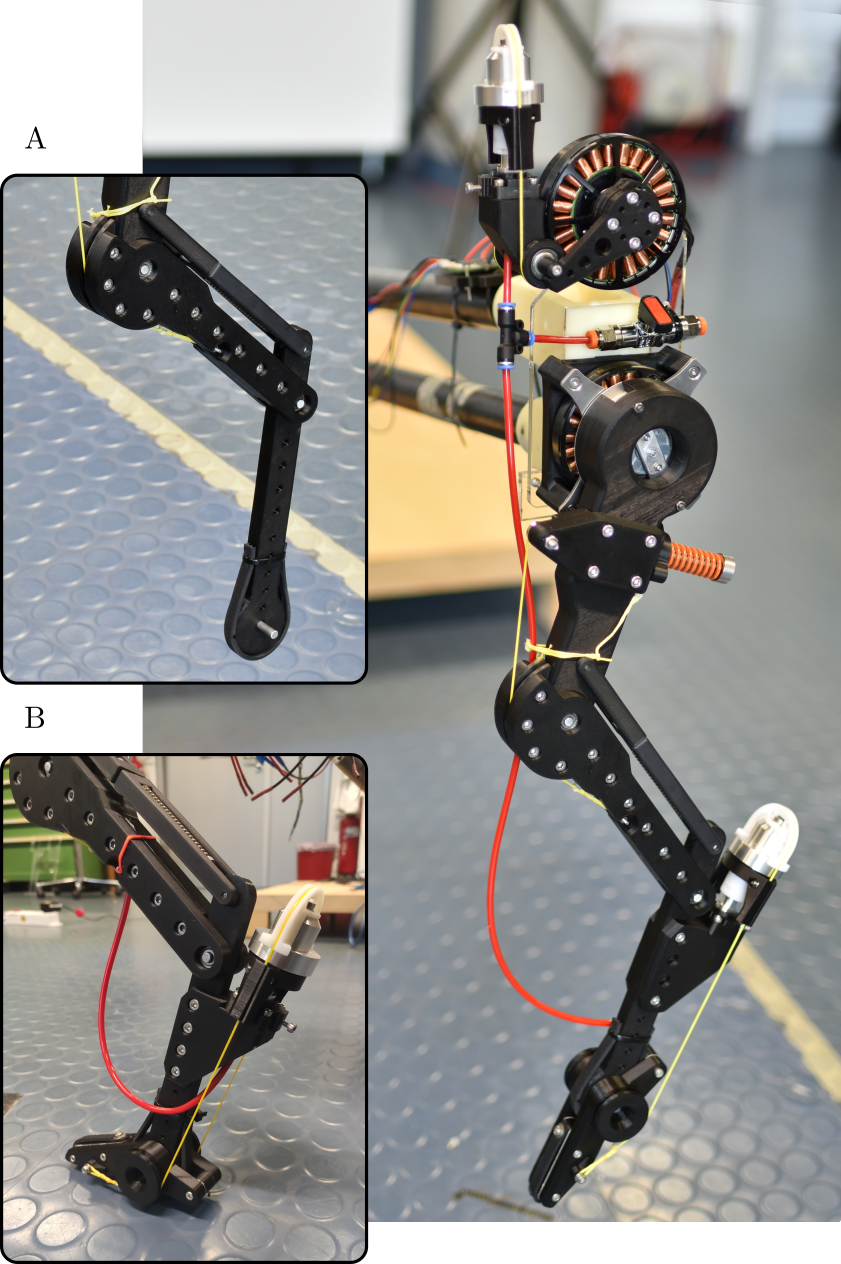}
    \caption{Experimental prototype of the bio-inspired leg. Detail figures on the left show the two configurations compared in this work. \textit{Configuration-A:} traditional bio-inspired layout \cite{ruppert_series_2019}. \textit{Configuration-B:} novel configuration with `foot' segment remotely actuated by a compliant pneumatic transmission.}
    \label{fig:photo}
\end{figure}

\section{Introduction}
We are interested in developing and characterizing legged robots with distal actuation; i.e., placing the ankle actuator into the robot's trunk, while transmitting power remotely, efficiently, and in an uncoupled manner. Placing heavy actuators close to the trunk leads to low-mass and low-inertia legs, which are well suited for agile and dynamic locomotion, simplifies legged robot control \cite{mochon1980ballistic}.

Control and design complexity increase when actuation is transferred over multiple joints, as is the case with robotic legs that feature more than two actuated segments. Multi-joint transmission can be achieved with cable (tendon), chain, and belt systems \cite{grizzle_mabel_2009,sprowitz_towards_2013,nejadfard_design_2019}. Simple configurations lead to mechanical coupling, where the output of one motor affects the movement of several, in-between connected joints. Decoupling mechanisms is either based on sensory feedback and kinematic control or complex mechanical solutions \cite{grizzle_mabel_2009,kim2017anthropomorphic}.

Remote actuation bypasses joints by transmitting power directly into the target location; Bowden cable transmissions and fluidic transmissions (hydraulic or pneumatic) are two common examples~\cite{narioka_development_2012}. Bowden cable power transfer is well tested, especially for lower frequency applications \cite{rutishauser_passive_2008}. However, at higher frequency gaits above a locomotion frequency of $f=\SI{3}{Hz}$, high friction between the cable core and its shaft leads to power losses that are prohibitive for mobile actuation solutions. Alternatively, industrial hydraulic piston actuators can be used in legged robots~\cite{raibert_trotting_1990,semini_design_2011}. Hydraulic piston designs are a compromise between their maximum working pressure and the fit between the seal, the piston, and the cylinder. With high input pressure, they can produce high-output forces and power. However, high-power piston-actuation requires tight seal fits that will increase stick and sliding friction and power losses, whereas low-friction piston-actuation reduces maximum working pressure \cite{johannesson_optimization_1980,nabae2018super}.

Rolling diaphragm actuators \cite{whitney_low_friction_2014} have the potential for remote and non-coupled actuation of distal robot leg and arm joints. Rolling diaphragm actuation features low friction ($<\SI{0.8}{\%}$ of the rated torque), which improves efficiency and leads to transparent, high-stiffness, backdrivable transmission, and simpler control \cite{bolignari2020design}. The transparent transmission allows for precise torque control, which provides a better base for advanced controller design. Its promising characteristics lead to robotic applications such as MRI compatible robots \cite{burkhard_rolling}, fluid dampers \cite{mo_effective_2020}, bio-inspired robots \cite{hepp_novel_2021}, and backdrivable manipulators and haptic devices \cite{whitney_hybrid_2016,frishman2021extending}.
In animals, both proximal and distal leg joints are actuated by muscle-tendons that can be presented as series elastic actuators \cite{alexander_three_1990}. Notably, distal muscle-tendon structures tend to store more elastic energy than proximal structures \cite{alexander_1977_storage}. Animals' locomotion shows how agile, robust, and energy-efficient these creatures can run, jump, and hop based on series elastically actuated structures embedded in their multi-segment legs. Animals feature low mass and moment of inertia at distal legs, with heavy actuators (muscles) mounted proximally \cite{witte_biomimetic_2004}. The exact functionality of the segment architecture, details of muscle-tendon units, and their mechanical and control coupling are not yet understood~\cite{mussaivaldi_motor_2000,fischer_basic_2002,cruse_insect_2007,ijspeert_central_2008,kano_tegotae_2017}. With that in mind,  actuator developments like the here presented series elastic diaphragm for distal actuation will allow building legged robots that can mimic complex animal leg structures, and test mechanical and control blueprints informed by animals.

In human walking, much research focuses on ankle kinematics and dynamics, the coupling of proximal and distal leg joints through elastic structures and control \cite{ivanenko_five_2004,de_groote_task_2014}, and the resulting impact on locomotion efficiency, agility, and robustness \cite{caputo_prosthetic_2015,sawicki_pneumatically_2009,farah_power_2015}. Related, catapult-like power output rooted in lower-leg muscle-tendon structures has been observed that exceeds the muscle's power output, by charging and discharging lower-leg series elasticities favourably \cite{tendon_2005,lipfert_impulsive_2014}. These examples from legged locomotion biomechanics emphasize the potential to reduce energetic losses, simplify control and mechanics, efficiently make use of small-sized and lightweight actuators, and increase robustness and agility based on distally acting series elastic actuators in legged machines.

In robots, distal leg joint actuation is often avoided. Powerful motor gearbox combinations come with a high mass and high moment of inertia, and mechanical complexity. High ankle output power is possible in stance phase and can be combined with mechanical series elasticity \cite{cherelle_amp_foot_2016,omineshita_robotic_2019}. But the actuators' mass is energetically costly to accelerate and decelerate through leg swing. Large distal masses tend to mechanically wobble from leg impact at the transition from swing to stance, which complicates sensing and control \cite{schmitt_human_2011}. Pneumatic artificial muscles (PAMs) are well-working, lightweight examples for remote actuation with high force and power output \cite{niiyama_mowgli_2007}. However, PAMs are actuated against the strain of their rubber-like bladder actuator, which decreases efficiency and power output, and increases control complexity~\cite{li_analysis_2020}.

In this project, we developed a \textbf{S}eries \textbf{EL}astic \textbf{D}iaphragm for distal \textbf{A}ctuation (SELDA). SELDA is inspired by the omnipresent series elastic actuation (SEA), especially in the distal animal leg \cite{alexander_three_1990}. With SELDA, we aim to develop hardware for agile legged hopping that is easy to control, is uncoupled from neighboring joint's movements and loads, easy and flexible to mount, features a remote motor placement and a distal power output, is distally lightweight, inherently compliant, and mechanically efficient. 

To test our design, we built a lightweight bio-inspired leg with a remotely actuated foot segment. The foot's actuator is placed in the robot's torso and its torque is reflected at the foot joint using a pneumatic rolling-diaphragm transmission, with compliance characteristics well suited for legged hopping. Our results also present an investigation into foot segments for bio-inspired robotic legs. We present experiment-based characterizations of the pneumatic transmission and actuation, and a comparison of hopping performances between two leg configurations (\Cref{fig:photo}); 
configuration-A is a leg without a foot segment, similar to previous work \cite{ruppert_series_2019}. Configuration-B is a leg with a foot segment, actuated by SELDA. 

In the following sections, we present the mechanical design and controller details of SELDA, our series elastic diaphragm for ankle actuation. We compare leg configurations A and B and record locomotion data from both robot legs when hopping in a circle, fixed to a rotating boom. Configuration-B is tested a) in passive mode, i.e., without foot-motor actuation to assess its compliance features, and b) when actively controlled, to explore the effects of foot-actuation timing during the step cycle.
\section{Robot Design}
The robot leg with its SELDA-actuated ankle weighs around \SI{1.2}{kg}. It features two motors; the hip motor drives the leg's femur segment and swings the leg forward and backward, with a trunk rigidly connected into the boom's base. This is the robot leg's main actuation. The ankle motor is proximally mounted and part of the distal SELDA actuation. The leg features four segments; three leg segments in a pantograph configuration similar to \cite{ruppert_series_2019}, and an additional foot segment of \SI{70}{mm} length. The robot's knee joint is passively extended by its knee spring acting on a knee cam. A biarticular spring mounted into the parallel segment acts over two joints. The compliant, under-actuated robot leg stores elastic energy in its mechanical springs during deceleration and spring energy is converted back to kinetic energy which accelerates the robot in the second half of stance phase. Effectively, the hip actuator is acting largely on the robot leg's virtual leg angle, which is the virtual line between the connection hip axis and foot axis. We expect the SELDA ankle actuator to input mechanical power both during forward and vertical motion.

\begin{figure}[t]
 \centering
 \includegraphics[width=\linewidth]{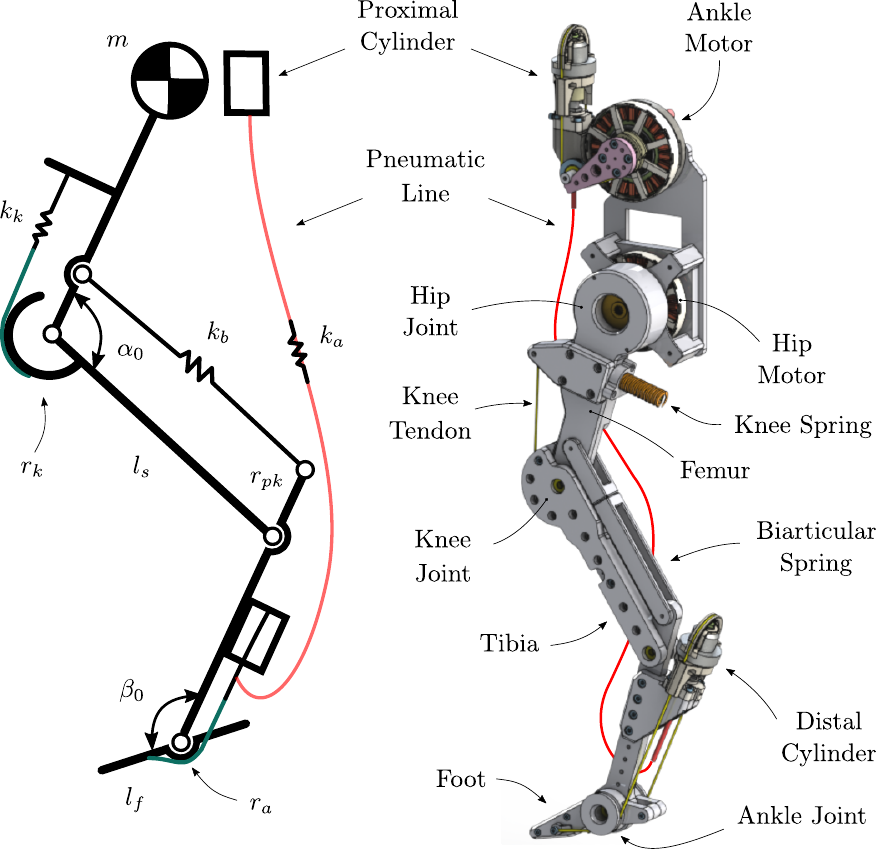}
 \caption{Left: schematic presentation of the experimental prototype. Right: computer-aided design (CAD) picture with the ankle joint mounted at the trunk above the hip joint, and connected through a hose (red) to the distal cylinder. A tendon couples the actuator's output to the ankle joint. The robot's second actuator is the hip motor which drives the femur segment.}
 \label{fig:drawing}
\end{figure}

In this work, the robot leg is minimally instrumented as we are aiming at comparing two robot configurations---with and without actuated foot---which we control with feedforward reference trajectories. Our goal is a setup that allow comparing leg configurations and their locomotion patterns fairly.
Observing biomechanics research with actuated orthosis we find that ankle actuation and especially push-off have an immediate and strong effect on the locomotion patterns \cite{sawicki_pneumatically_2009,caputo_prosthetic_2015}. Consequently, tuning parameters for high-power ankle actuators requires advanced controllers and high-bandwidth feedback signals. With our goal in mind of developing SELDA and comparing it to a previous design, we under-dimensioned the ankle joint's output torque by limiting the ankle's cam size. Because ground reaction forces drop between mid-stance and lift-off to zero, even smaller ankle torques from active ankle actuation will induce leg extension forces when met with equilibrium conditions, and we expected to observe effective changes in forward speed and hopping height.

\subsection{Diaphragm pneumatic transmission}
\begin{figure}[t]
    \centering
    \includegraphics[width=\linewidth]{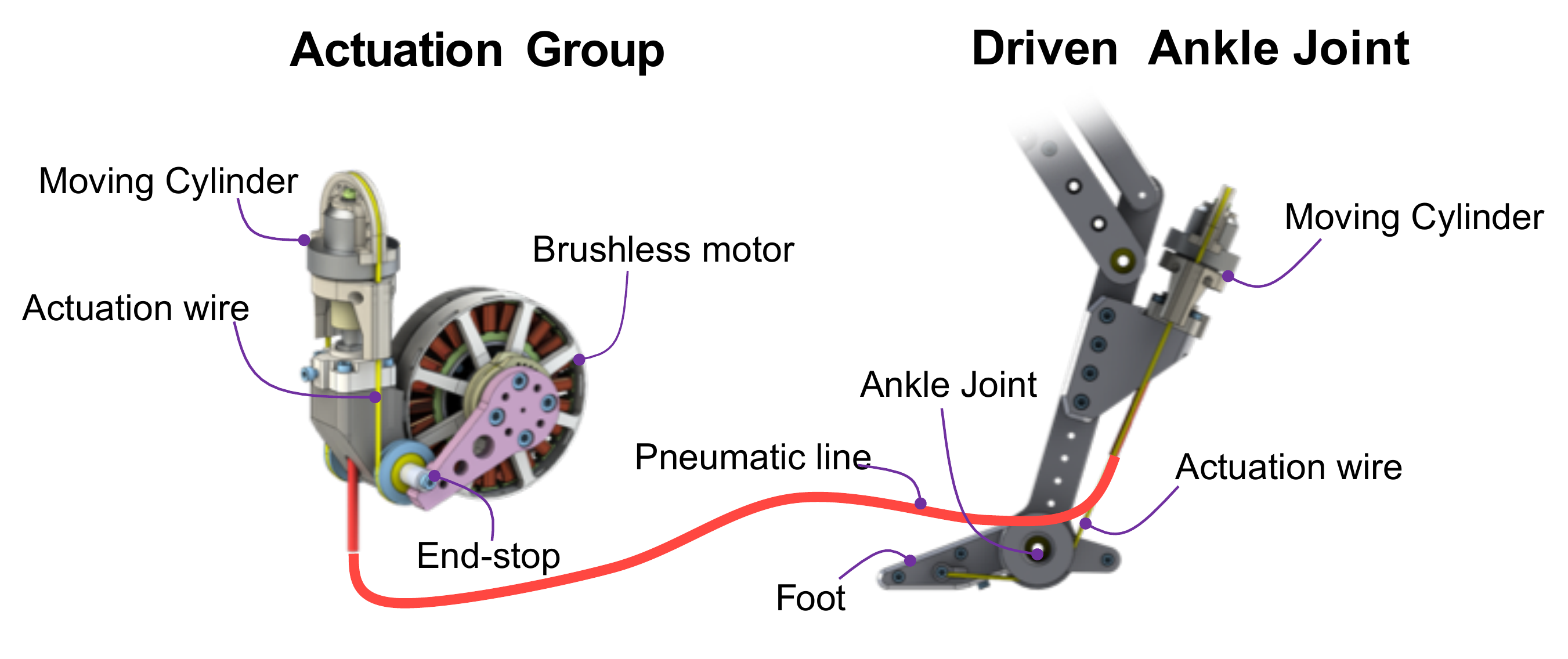}
    \caption{CAD picture of the SELDA rolling diaphragm transmission actuating the ankle joint remotely.}
    \label{fig:plot:hydrotransmission}
\end{figure}

In our design, we employ the concept of rolling diaphragm hydrostatic transmission \cite{bolignari2020design} to implement a series elastic diaphragm for distal actuation (SELDA) which integrates the functionalities of remote actuation and series elastic actuation in a single lightweight, efficient and compact structure. 
SELDA employs hydraulic cylinders featuring rubber-made diaphragm sealing elements that provide an effective mechanism for transmitting torques and forces over distance showing attributes that include a high-force/torque friction/backlash-free operation, high bandwidth, easy to control, and relatively low cost. 
In our design, to minimize weight and to simplify the structure, we take advantage of the asymmetrical torque requirements for torque at the ankle joint, i.e., large torque is required in one direction to generate large downward force during the pull off the ground but very small return torque is needed in the other direction to recall on the foot during the swing phase. According to this requirement, we implement a pneumatic SELDA equipped with a single fluid line  operated by a cylinder with fixed-piston (i.e., moving cylinder \cite{bolignari2020design}) that transmits the action of a brushless actuator combined with a planetary gear (located on the body) in the direction of foot pull-off (see \Cref{fig:plot:hydrotransmission}). The required return movement of the foot is simply obtained taking advantage of the offset force produced by a pre-pressurization of the line that acts against an end-stop of the motor rotation. 
A pneumatic line of a specific length and diameter is chosen to provide a calibrated stiffness of the transmission which results in a series elastic element embedded in the distal joint transmission. 

\subsection{Bio-inspired leg design and details}
We implemented our hydrostatic transmission on a bio-inspired complaint robotic leg (\Cref{fig:photo}). Inspired from \cite{ruppert_series_2019}, this leg include four segments and three passive elastic elements that mimic the compliant behavior of mammalian quadruped's leg.
The hip and the ankle joints are actuated by means of brushless motors (model NM7005 KV115 by \textit{T-Motor}, with $\SI{1.3}{Nm}$ maximum rated torque). The hip motor is equipped with a 5:1 planetary gearbox (model RS3505S, \textit{Matex}). The motor positions are measured by rotary encoders (model AEAT8800-Q24, \textit{Broadcom}, \SI{12}{bit})).
We use open-source drivers (Micro-Driver, \cite{Grimminger_2020}) for motor control, current sensing, and encoder reading. The Micro-Driver board is capable of dual motor Field Oriented Control (FOC) at \SI{10}{kHz}. We implemented our controller on a single board computer (3B+, \textit{Raspberry Pi Foundation}) with control frequency of \SI{1}{kHz}.
The proximal module for ankle actuation is mounted above the hip and it is connected to the distal diaphragm cylinder through a \SI{5}{mm} polyurethane pneumatic hose.
The knee joint is coupled to the knee spring (model SWS14.5-45, \textit{MISUMI}, $k_\mathrm{k} = \SI{10.9}{N/mm}$) that extends the knee to resting angle of $\alpha_0=\SI{130}{\degree}$. In parallel to the shank segment, a spring-loaded (model UBB10-60, \textit{MISUMI}, $k_\mathrm{b} = \SI{9.8}{N/mm}$) biarticular segment replicates a pantograph kinematics.
This segment replicates the lower leg muscle-tendon apparatus of gastrocnemius muscle and Achilles tendon. Details of design parameters can be found in \Cref{tab:design_para}.

\begin{figure}[t]
    \centering
    \includegraphics[width=\linewidth]{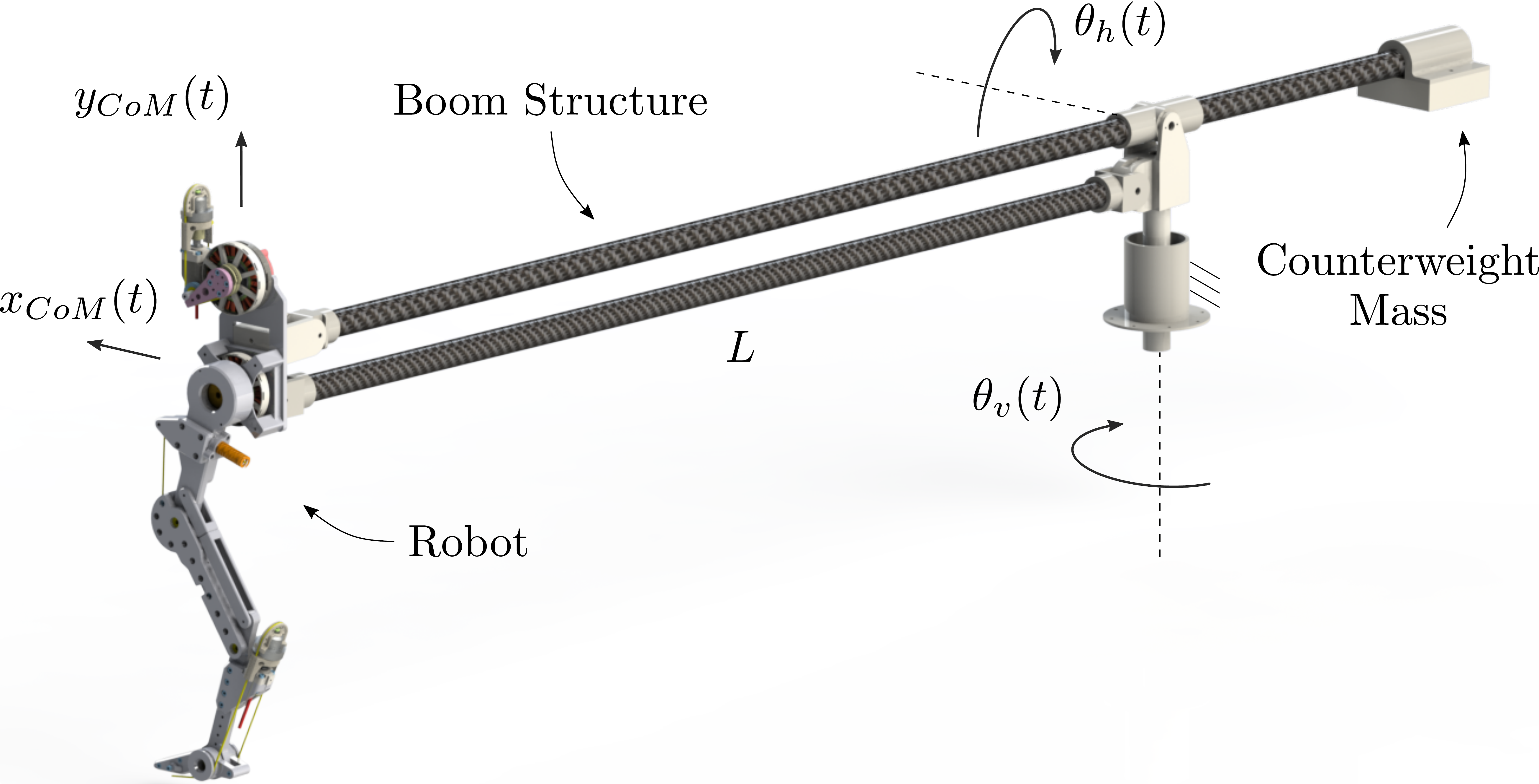}
    \caption{Experimental test bench. The leg is constrained to a boom structure to allow horizontal and vertical displacement only; torso rotation and lateral displacement are prevented. This configuration allows the robot to travel long distances.}
    \label{fig:boom}
\end{figure}

The hip of the robot is mounted to the boom structure (\Cref{fig:boom}), which allows the robot to jump along a circular path over long distances.
The boom prevents torso rotation, eliminating the need for trunk pitch control.
The length of the boom rods $L$ is \SI{1.55}{m} determining a travelled distance of \SI{9.73}{m} over one entire revolution. 
We connected a counterweight at the opposite end of the boom that balances the own mass of the boom, so that the robot does not undergo additional weight.
The boom rotation angles $\theta_h$ and $\theta_v$ are measured by two rotary encoders (model 102-V, \textit{AMT},  \SI{11}{bit}).
The position of the robot center of mass (CoM) $x_{CoM}$ and $y_{CoM}$ are evaluated as function of $\theta_h$ and $\theta_v$ angles.%
\begin{table}[!h]
\caption{Robot design parameters. Adding the SELDA actuated foot segment increases the robot's weight by \SI{0.15}{kg}}
\vspace{-4mm}
\label{tab:design_para}
\begin{center}
\begin{tabular}{lcl}
Parameters 						& 		& Value \\
\hline
Robot mass configuration A  	& \textit{$m_A$}		& \SI{1.05}{kg} \\ 
Robot mass configuration B  	& \textit{$m_B$}		& \SI{1.20}{kg} \\ 
Segment length					& \textit{$l_s$}		& \SI{150}{mm} \\
Foot length						& \textit{$l_f$}		& \SI{70}{mm} \\
Knee spring stiffness 			& \textit{$k_k$}		& \SI{10.9}{N/mm} \\
Bi-articular spring stiffness  	& \textit{$k_\mathrm{b}$}	& \SI{9.8}{N/mm} \\
Ankle stiffness	(air spring)	& \textit{$k_a$}		& \SI{0.15}{Nm/rad} \\
Knee spring pulley radius		& \textit{$r_k$} 		& \SI{30}{mm} \\
Bi-articular insertion radius	& \textit{$r_{pk}$}		& \SI{30}{mm} \\
Ankle pulley radius				& \textit{$r_a$} 		& \ang{30} \\
Knee resting angle				& \textit{$\alpha_0$} 	& \ang{130} \\
Ankle resting angle				& \textit{$\beta_0$} 	& \ang{160} \\
Leg resting length				& \textit{$l_0$}		& \SI{408}{mm} \\
Hip swing amplitude             & \textit{$A$}          & \ang{18}  \\
Hip oscillation frequency       & \textit{$f$}          & \SI{1.65}{Hz}  \\
\end{tabular}
\end{center}
\vspace{-4mm}
\end{table}
%
\section{Transmission Characterization}
\begin{figure}[t]
    \centering
    \includegraphics[width=\linewidth]{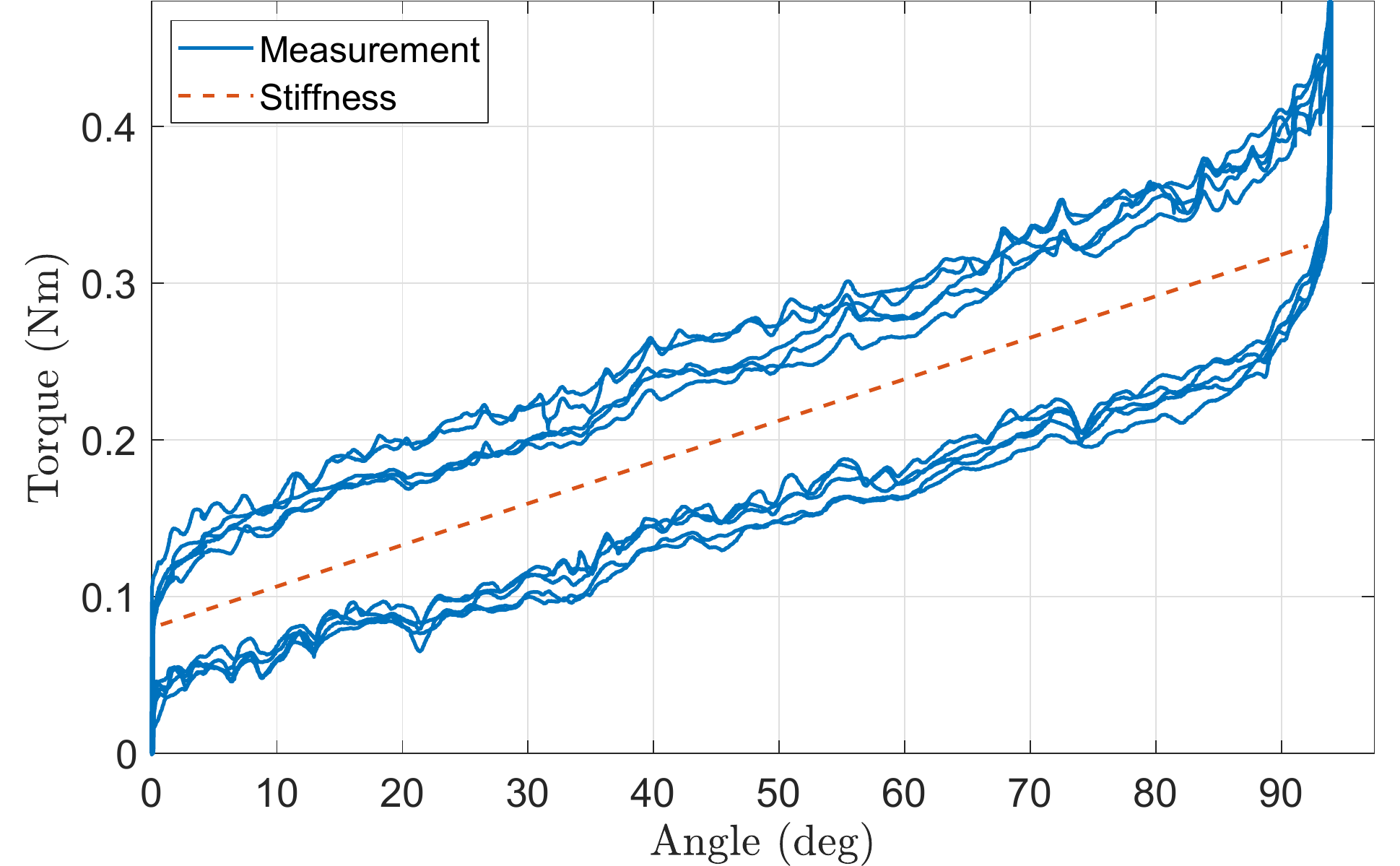}
    \caption{Stiffness characterization of the transmission system. Estimated stiffness of 0.15 Nm/rad from the proximal cylinder side.}
    \label{fig:stiffness}
\end{figure}

The air spring feature of the transmission adds compliance to the distal joint. Compliance is therefore characterized in terms of transmission stiffness in \Cref{fig:stiffness}.
The transmission is pre-pressurized to \SI{0.5}{bar}, the foot is consequently fully extended, and a full-stroke rotation is applied to the motor. %
We manually rotated the rotor with an instrumented lever arm equipped with a load cell (model 3133\_0, \textit{Phidgets}) at its extremity for accurately measuring the applied torque. We measured a transmission stiffness of \SI{0.15}{Nm/rad}.

\section{Hopping Experiments} \label{sect:hopping:experiments}
\begin{figure*}[!h]
\centering
\includegraphics[scale=1]{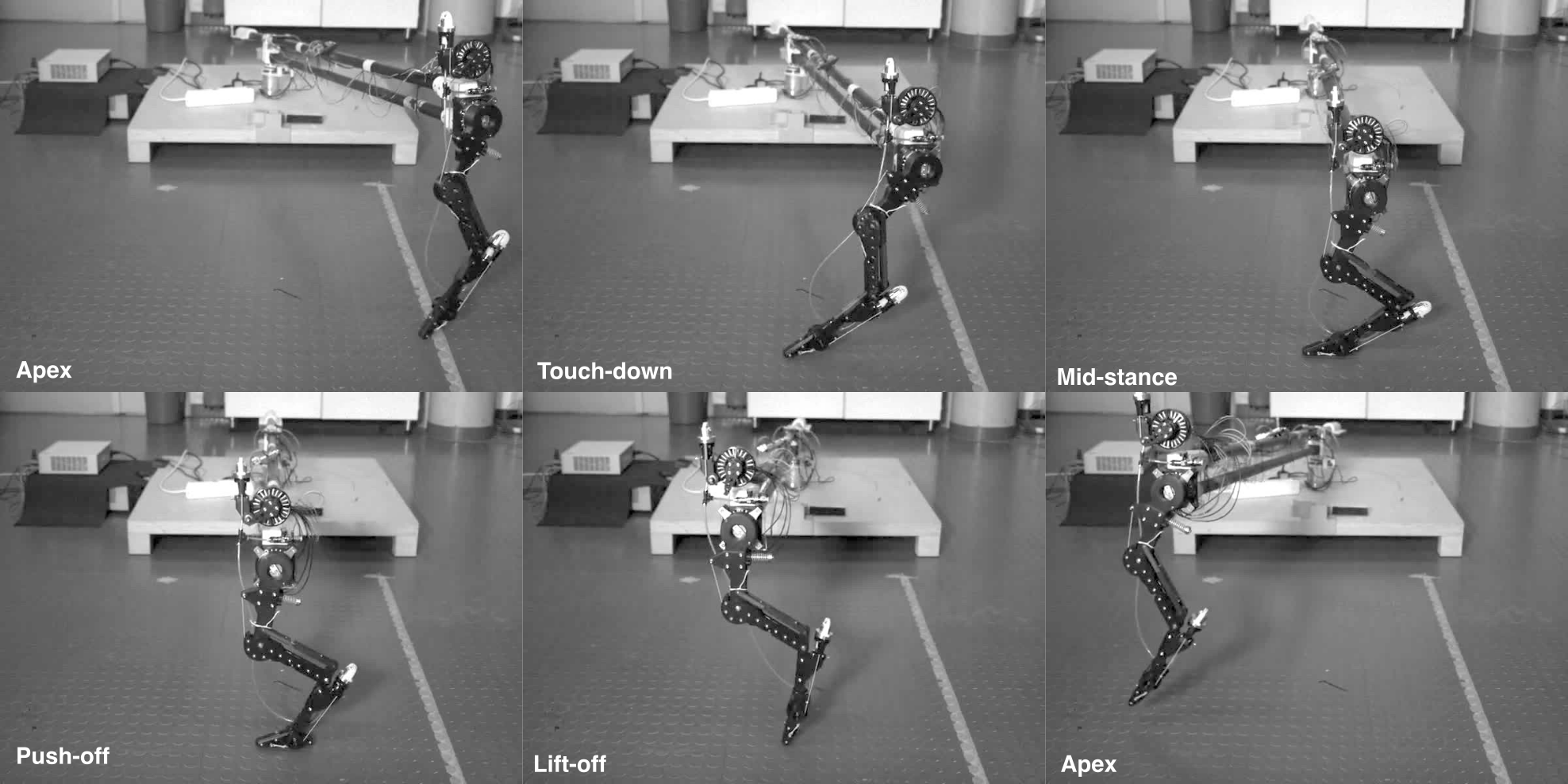}
\caption{Gait cycle snapshots from apex to apex, taken from high-speed video footage. A \emph{delay} was programmed to trigger push-off actuation. Step cycle time ($T=1/f$, $f$ is frequency) is \SI{606}{ms}, or around \SI{1.65}{Hz} hopping frequency.}
\label{fig:snapshots}
\end{figure*}
This section describes the hopping experiments. \Cref{sect:hopping:config} described the experimental configuration chosen for gait analysis; \Cref{sect:hopping:passive} compares the performance of the leg with and without the distal segment, i.e. configurations-A and -B in \Cref{fig:drawing} respectively; \Cref{sect:hopping:active} presents our first investigation of the effect of distal actuation, mainly focusing on the kinematic effects in terms of gait velocity, step length and step height. A typical gait obtained during the experiments with SELDA system is shown in \Cref{fig:snapshots}.
\subsection{Experimental Configuration} \label{sect:hopping:config}
\begin{figure}[t]
    \centering
    \includegraphics[width=\linewidth]{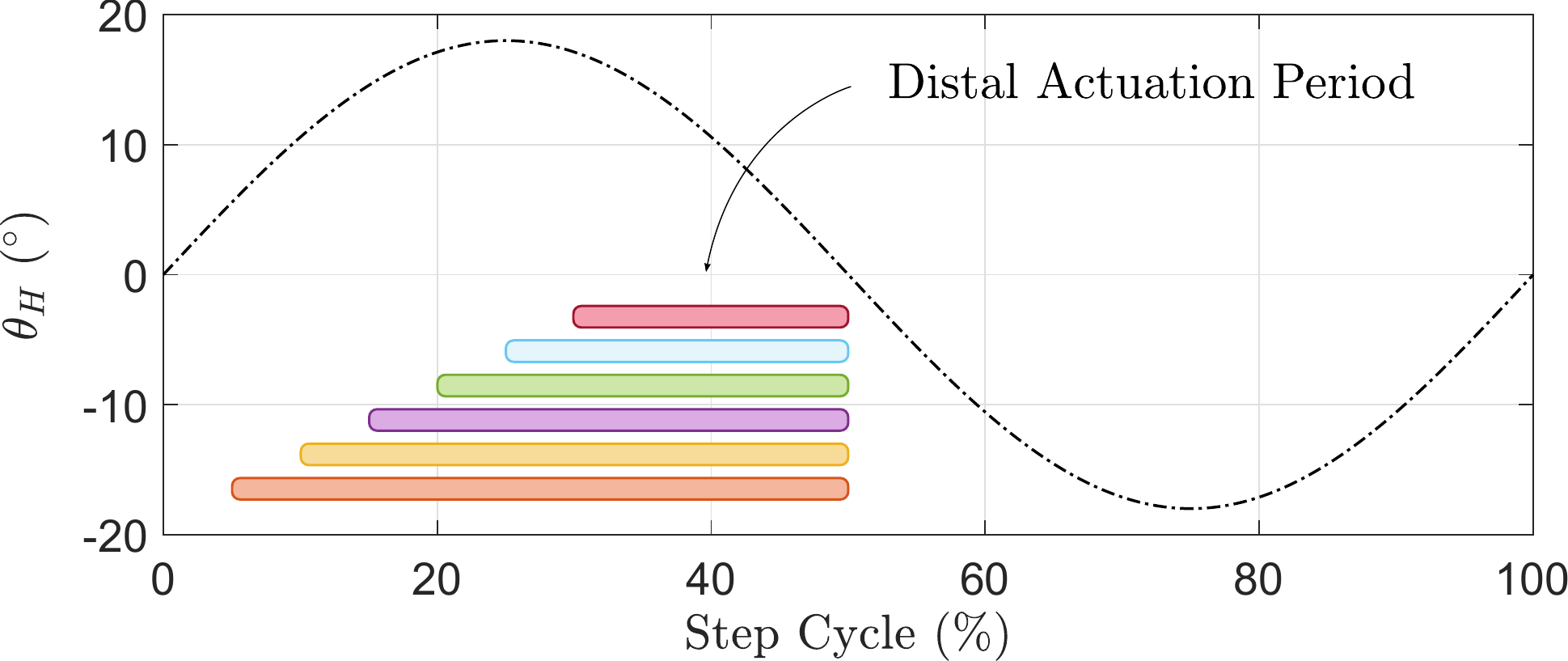}
    \caption{Distal actuation timing with respect to the hip reference trajectory (black dashed line). The colored rectangles indicate the period of distal actuation. Different initial timing $t_T$ values are considered: $t_T = 5, 10, 15, 20, 25, 30 \%$ of the step cycle; distal actuation is then ended at \SI{50}{\%} of the step cycle, during the swing phase. According to scheme in \Cref{fig:drawing}, positive angles indicate a leg position behind the vertical axis.}
    \label{fig:timing}
\end{figure}

The robot response is characterized in case of simple control strategies in order to emphasize intrinsic self-stabilizing response through compliant design. The hip joint is position controlled along a sinusoidal trajectory:
\begin{equation} \label{eq:sine}
    \hat{\theta}_h = A sin( 2 \pi f )
\end{equation}
where constant $A$ is the hip trajectory amplitude and constant $f$ defines the hopping frequency.
Hip oscillation amplitude $A = 18^\circ$ and locomotion frequency $f$ = \SI{1.65}{Hz} are common for all of our experiments.
Note that the chosen parameter set is likely not optimal for both configurations; we expect that each configuration has its dynamics. 
Nevertheless, we keep the parameter common for a consistent comparison between leg configurations. 
An example hip trajectory is shown in \Cref{fig:timing}.
The trajectory tracking is performed through a PD controller:
\begin{equation} \label{eq:controller}
    \tau_h = k_p e(t) + k_d \dot{e}(t)
\end{equation}
where $\tau_h$ is the commanded torque to the hip joint 
and variable $e(t)$ is the tracking error $e(t) = \hat{\theta}_h - \theta_{h}$.
The controller \Cref{eq:controller} behaves like a virtual spring-damper element acting between the reference trajectory $\hat{\theta}_h$ and the hip joint with stiffness value $k_p$ and damping coefficient $k_d$. Parameters $k_p = 40$ and $k_d = 0.35$ are fixed for all experiments.

To investigates the influence of the distal actuation, we focus on the actuation timing of the foot segment during the step cycle.
A step torque reference of \SI{1}{Nm} is commanded to the ankle motor: the initial actuation instant varies in the range \SIrange{5}{30}{\%} of the step cycle, as shown by the colored bars in \Cref{fig:timing}; the  ankle actuation is then ended at $50\%$ of the step cycle, during the swing phase when the leg is not in contact with the ground.
The compliance features of the pneumatic transmission allow driving a simple step torque-reference to the ankle motor so that we can focus on analyzing the influence of actuation timing only. 

\begin{figure}[t]
    \centering
    \includegraphics[width=\linewidth]{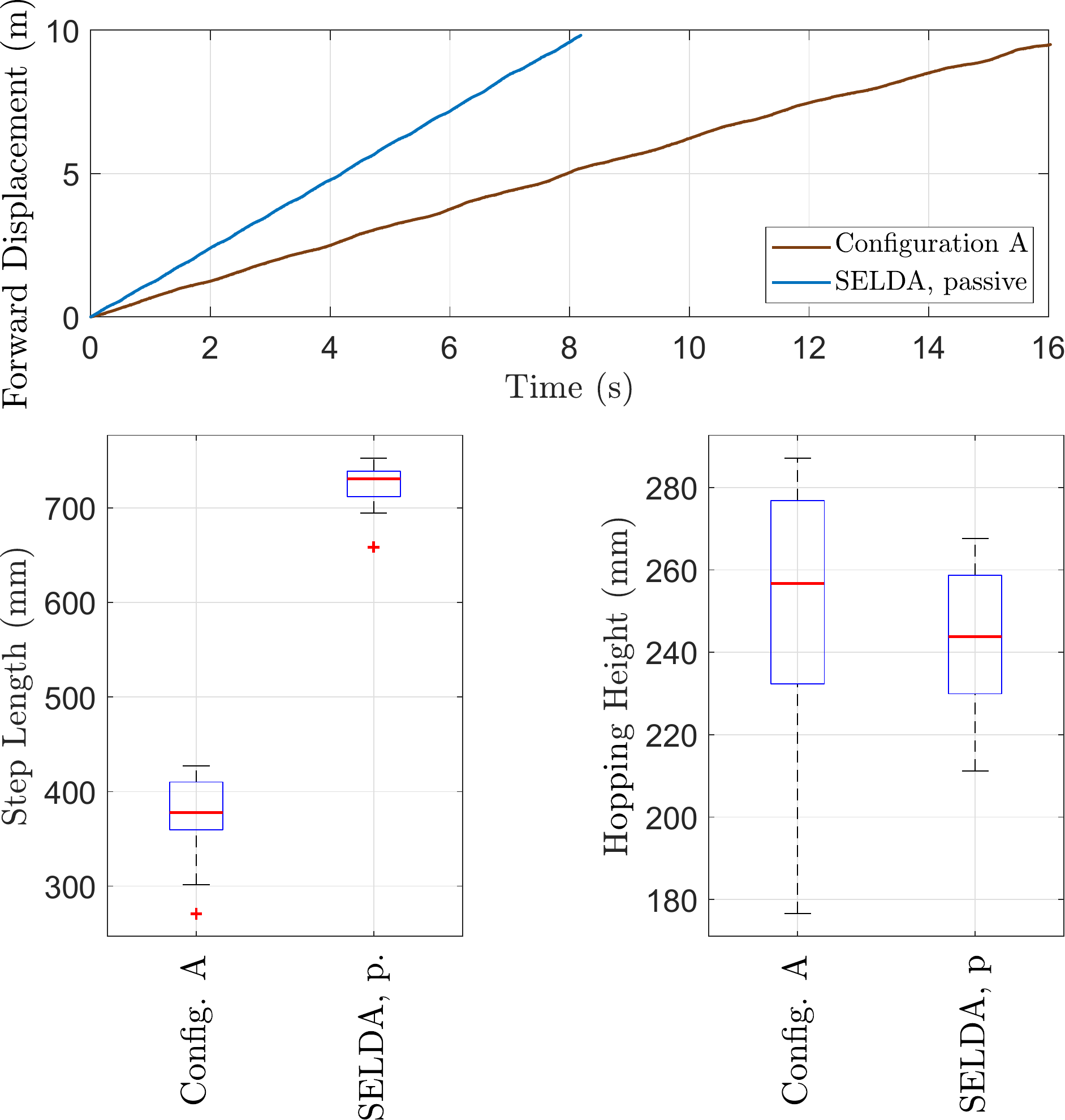}
    \caption{Performance comparison between configurations-A and -B, (\Cref{fig:drawing}); in this experiment the distal segment is not actuated and the pneumatic transmission behaves like an air spring. \textit{Top:} time taken to travel a complete turn around the boom. \textit{Left:} statistical analysis of the maximum step height over one full boom revolution. \textit{Right:} statistical analysis of the step length over an entire revolution.}
    \label{fig:plot:noToe}
\end{figure}

\begin{figure*}[t]
    \centering
    \includegraphics[width=\linewidth]{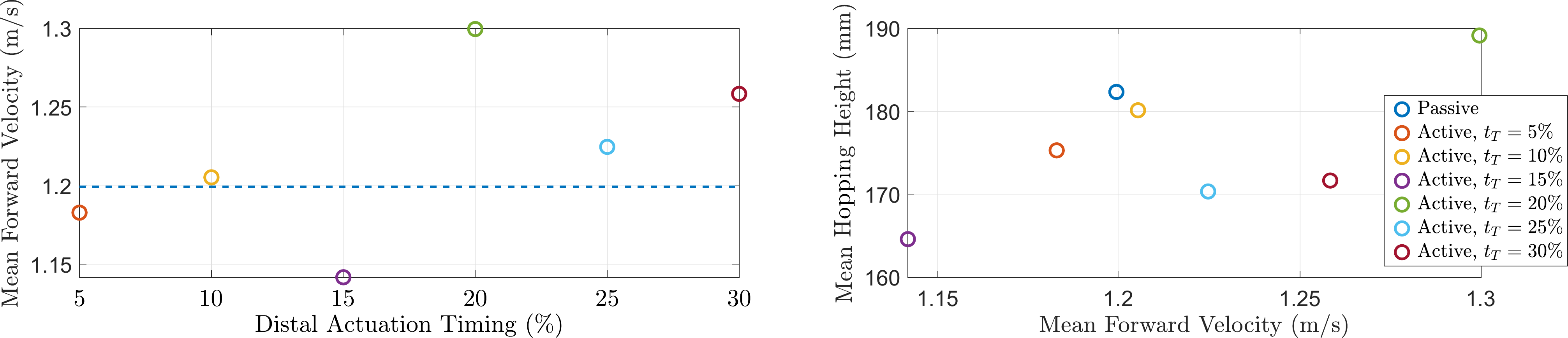}
    \caption{Investigation of the effect of different distal actuation timing $t_T$. \textit{Left:} mean velocity of the center of mass achieved for different values of timing $t_T$. \textit{Right:} mean step height with respect to mean forward velocity.
    Step height is the difference between the highest and the lowest vertical position of the robot's center of mass at each step.
    Mean values refer to the dataset corresponding to one full revolution around the boom (\SI{9.7}{m} travelled distance) during steady state locomotion.}
    \label{fig:plot:activeVel}
\end{figure*}

\begin{figure*}[t]
    \centering
    \includegraphics[width=\linewidth]{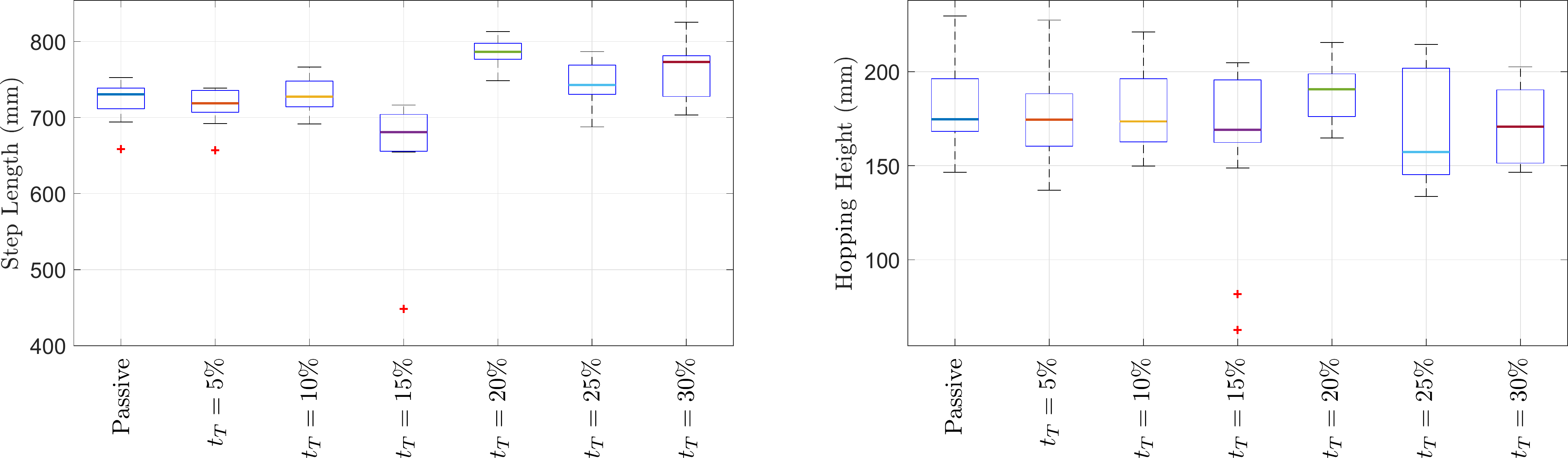}
    \caption{Box plot statistical representation of the influence of distal actuation timing $t_T$ on step length and maximum step height. The dataset of this analysis corresponds to samples collected during one full revolution around the boom in steady state gait. The minimum number of steps for one revolution is 13.}
    \label{fig:plot:activeBox}
\end{figure*}

\subsection{Analysis of Passive Foot} \label{sect:hopping:passive}
In this section, leg configuration-B is tested in passive mode, i.e., without activating the ankle motor, and its performance is compared to configuration-A in order to assess the benefits of the additional compliant foot.
\Cref{fig:plot:noToe} shows this comparison evaluated on data sets corresponding to a complete revolution of the circular trajectory around the boom. The top plot shows that the distance of \SI{9.7}{m} is travelled in \SI{15.7}{s} by robot configuration-A and in \SI{8.1}{s} by the configuration-B in SELDA passive mode.
In this experiment, SELDA increases the forward velocity $\dot{x}_{CoM}$ from \SIrange{0.62}{1.20}{m/s}; an almost two-fold increase.
The higher-speed locomotion is also visible in terms of step length, \Cref{fig:plot:noToe}~(bottom-left), which increases of \SI{93}{\%} from \SI{378}{mm} to \SI{730}{mm}.
The foot slightly affects the robot's maximum hopping height (\Cref{fig:plot:noToe}, bottom-right), but it leads to a more repeatable hopping height and more stable hopping motion.

\subsection{Analysis of Active Foot} \label{sect:hopping:active}
This section investigates the effect of the SELDA activation timing, also in comparison with the passive foot configuration.
The initial timing $t_T$ of the ankle actuation varies in the range \SIrange{5}{30}{\%} of the step cycle; a timing of \SI{5}{\%} means that the ankle is actuated right after the touch-down, while a timing of \SI{30}{\%} means that the ankle is actuated right before the lift-off.
\Cref{fig:plot:activeVel}~(left) shows that, in general, the center-of-mass velocity $\dot{x}_{CoM}$ can be increased by activating the ankle after mid-stance, while it is slowed down by actuating the ankle before mid-stance. In particular, the active SELDA achieves the highest forward velocity of \SI{1.30}{m/s} with an actuation timing of $t_T =$ \SI{20}{\%} versus a speed of $\dot{x}_{CoM} =$ \SI{1.20}{m/s} for the passive SELDA. The lowest performance is observed with an actuation timing of $t_T = \SI{15}{\%}$ leading to an average forward speed of $\dot{x}_{CoM} = \SI{1.14}{m/s}$.
\Cref{fig:plot:activeVel} (right) illustrates the energy transfer between hopping height and forward velocity $\dot{x}_{CoM}$. Hopping height is calculated at each step as the difference between the highest and the lowest vertical position $y_{CoM}$ reached by the robot's center of mass.
By tuning the activation timing, we can effectively adjust the hopping height by \SI{11}{\%} and the forward velocity by \SI{14}{\%}.
Note that our diaphragm actuation produces a torque of ($\approx$\SI{1}{Nm}) from the motor side, which also compensates for the internal pressure of the pneumatic line. Albeit the limited actuator output torque, we observe that locomotion speed and hopping height are effectively altered (\Cref{fig:plot:activeVel}, left and right).

We quantify the activation timing effect with step length and hopping height over steps (\Cref{fig:plot:activeBox}).
Narrow bands in the box plot indicates more stable hopping gait.
We observed period-2 hopping in some experiments, which expand the confidence internal in the plot.
If gait parameters such as frequency and amplitude are tuned to match the robot's own dynamic, we expect to further reduce variation of step length and hopping height between steps.

A video of the robot leg with SELDA actuated ankle joint can be found in the supplementary materials, and at \href{https://youtu.be/P1zkDmtD-hs}{this YouTube link}.

\section{Conclusions}
This work proposes distal actuation of the foot segment in a bio-inspired hopping robot with a compliant rolling diaphragm pneumatic transmission. 
Diaphragm actuation has appealing features of lightweight, low-friction, high efficiency, and truly remote actuation.
We found self-stabilizing gaits with comparatively simple, open-loop position control.
We show that the addition of the foot segment improves the locomotion performance of the robot, already in its passive elastic mode, with an increase in forward velocity of \SI{93}{\%}. 
With the actuated foot, we observed that actuation timing effectively influences the hopping gait.
By tuning the ankle actuation timing from \SI{5}{\%} to \SI{30}{\%} of the gait cycle, we observed \SI{11}{\%} change in hopping height and \SI{14}{\%} in forward velocity, with its currently under-dimensioned actuator. 
Based on these first results, our future developments will focus on providing the proximal cylinder with an antagonistic action to reduce the actuation effort that is required to balance the hose's internal pressure.
An additional gearbox mounted to the ankle motor will increase output torque at reduced electrical power requirements.
Note that the proposed experiments are not based on optimal control strategies and SELDA system may potentially achieve higher performances.
Gait patterns will be optimized for energy-efficient and agile locomotion, by tuning control parameters and type, and by introducing online feedback.

\bibliographystyle{IEEEtran}
\bibliography{main.bib}

\end{document}